\documentclass{article}
\usepackage{spconf,amsmath,graphicx}
\usepackage{booktabs}
\usepackage[export]{adjustbox}
\usepackage{amsmath,graphicx,bm}
\usepackage{multirow}
\usepackage{subfigure}
\usepackage{cite}
\usepackage{graphicx}

\title{TRAINING WAKE WORD DETECTION WITH SYNTHESIZED SPEECH DATA ON CONFUSION WORDS}
\name{Yan Jia$^{1}$, Zexin Cai$^{2}$, Murong Ma$^{1}$, Zeqing Zhao$^{3}$, Xuyang Wang$^{3}$, Junjie Wang$^{3}$, Ming Li$^1$}

\address{$^{1}$Data Science Research Center, Duke Kunshan University, Kunshan, China \\
 $^{2}$Department of Electrical and Computer Engineering, Duke University, Durham, USA \\
$^{3}$AI Lab of Lenovo Research, Beijing, China }

\begin{document}
\ninept
\maketitle
\begin{abstract}
Confusing-words are commonly encountered in real-life keyword spotting applications, which causes severe degradation of performance due to complex spoken terms and various kinds of words that sound similar to the predefined keywords. To enhance the wake word detection system's robustness on such scenarios, we investigate two data augmentation setups for training end-to-end KWS systems. One is involving the synthesized data from a multi-speaker speech synthesis system, and the other augmentation is performed by adding random noise to the acoustic feature. Experimental results show that augmentations help improve the system's robustness. Moreover, by augmenting the training set with the synthetic data generated by the multi-speaker text-to-speech system, we achieve a significant improvement regarding confusing words scenario.
\end{abstract}
\section{Introduction}
\label{sec:intro}

In intelligent speech processing applications, the keyword spotting(KWS) system, including wake-up word detection, plays an important role in human-computer communication. KWS aims to detect a predefined keyword or a set of keywords in a continuous audio stream. It is used to improve the interaction experience between human and machine, where the machine responds when it needs to and stays inactive otherwise.  

Studies have been proposed to deliver robust approaches with high detection accuracy, and to overcome issues met in practical applications regrading real-time response, computational cost and memory consumption. Alan and Robert adopted dynamic time warping (DTW) for keyword spotting back in 1985 \cite{1}. This template matching method is efficient yet the detection accuracy and robustness are highly limited by pre-selected templates. Other than matching in the acoustic feature level as in \cite{1}, approaches that use statistic modeling, e.g. hidden Markov models (HMM), for matching have shown better robustness \cite{2,3,4}. In addition, HMM involves modeling on the filler, which denotes all non-keywords, to improve the detection accuracy. 
The traditional Gaussian mixture models (GMM) were commonly used in statistic modeling for acoustic features in HMM-based approaches, and it is replaced by deep neural networks (DNN) recently, where significant improvement has been shown by adopting the latter architecture \cite{Sun2017CompressedTD, Panchapagesan2016MultiTaskLA}.  Alternately, end-to-end approaches were proposed to address the keyword spotting in a single model. Various neural network structures, including deep neural network (DNN) \cite{5}, convolutional neural network (CNN) \cite{Sainath2015ConvolutionalNN}, and recurrent neural networks (RNN) \cite{10.1007/978-3-540-74695-9_23,WOLLMER2013252}, have been applied as the backbone network of end-to-end approaches and shown marvelous performance in detecting wake-up words. 

However, unexpected problems occur when applying wake-up word detection in many practical applications. For example, the probability of false alarm becomes higher under complex acoustic environments and disambiguated content. Without further adaptation, the KWS system may misclassify filler as keywords since some of the filler actually sound close to the keywords.  Those are the confusion words that dramatically affect KWS performance. This issue hasn't been well addressed in end-to-end approaches, where detection results are directly drawn from the model. Moreover, it is costly to acquire adversarial samples for training a KWS system that accurately classifies confusion words.

In this paper, we proposed two augmentation techniques employed on an end-to-end approach to address the aforementioned issue. The idea is motivated by the maximum mutual information (MMI) criterion in the speech recognition area \cite{Povey+2016}, which is designed to improve the discriminative power of the model. The first technique is to augment the data by applying random masking on speech signals, and the second augmentation is performed by a text-to-speech (TTS) system. To the best of our knowledge, we are the first one to employ a multi-speaker speech synthesizer to generate adversarial samples of confusion words to improve the performance of the wakeup word detection system. Both augmentation methods achieve significant improvement on the end-to-end KWS model. Especially for TTS augmentation, the false alarm rate drops from 100\% to 0.083\% compared with the one that is evaluated with the system trained without augmentation. 

This rest of the paper is organized as follows. Section 2 describes the framework of the CNN based KWS system, and section 3 presents our augmentation methods. Section 4 discusses the experimental results, and the conclusion is provided in section 5.

\section{MODEL ARCHITECTURE}
\label{sec:format}
\begin{figure}[th]
    \centering
    \includegraphics[width=0.48\textwidth]{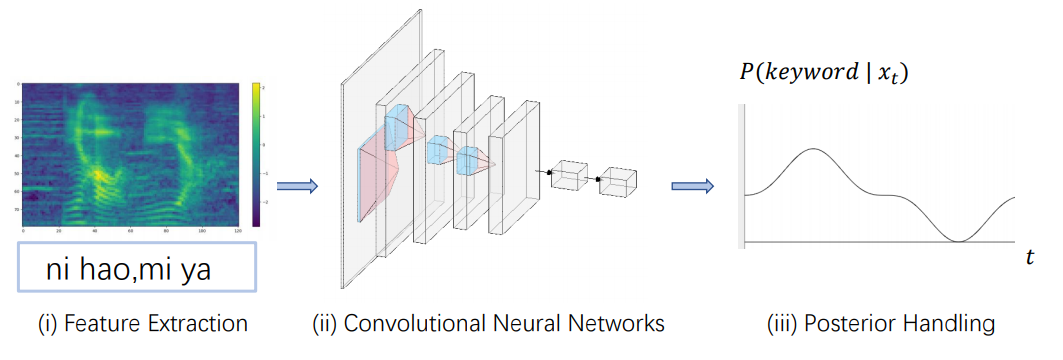}
    \caption{Framework of the baseline system.}
    \label{CNN_framework}
\end{figure}

In this section, we present our baseline system, which is modified from the CNN-based KWS system \cite{Sainath2015ConvolutionalNN}. As shown in Figure \ref{CNN_framework}, our baseline system consists of three modules:(i) a feature extraction module, (ii) a convolutional neural network and (iii) a posterior processing module. 

The feature extraction module converts the audio signals into acoustic features. 80 dimensional log-mel filterbank features are extracted for speech frame in 50ms window and 25ms frame shift. Then we apply a segmental window with 121 frames to generate training samples that contain enough context information as the input of the model. 

Our backbone network consists of three convolutional layers each followed by a maximum pooling layer.  For all three CNN layers, the kernel size is set to (3,3), the stride is (1,1), and the pooling size is set to (2,2). Two fully connected layers and a final softmax activation layer are applied as the back-end prediction module to obtain the keyword occurrence probability. 

The acoustic feature sequence is transformed into a posterior probability sequence of selected keywords by the model. We perform the keyword detection algorithm over a sliding window with length $T_s$. Here we use $\textbf{x}^{(i)} = \{x_i, x_{i+1}, \dots , x_{i+T_s} \}$ to denote one input window over the segment $X$ that contains $N$ frames. Then the keyword confidence score is calculated as follows:
\begin{equation}
    conf(X) = \max_{1\leq t\leq N-T_s} P_{keyword}(x^{(t)})
\end{equation}
where $P_{keyword}(x^{(t)})$ is the posterior probability of the keyword appearing in the window stated at frame $t$. The KWS system triggers once when the confidence score exceeds a predefined threshold.

\begin{figure}[t]
    \centering
    \includegraphics[width=0.3\textwidth]{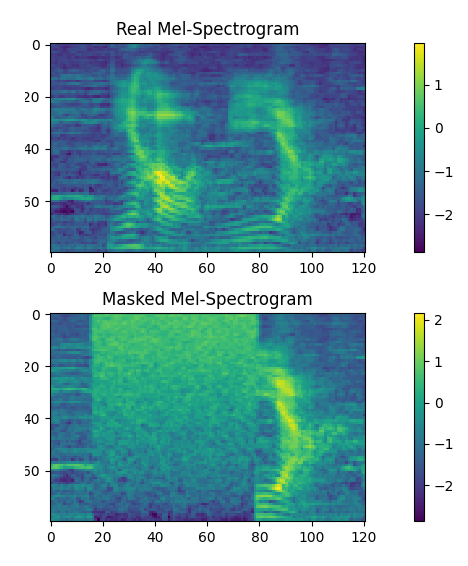}
    \caption{Real and Masked sample spectrum.}
    \label{Truemasked_figure}
\end{figure}

\section{Data augmentation}
The complexity of vocabulary and the speech condition matter a lot regarding the performance of speech-related applications, e.g., speech recognition and wake-up word recognition, in real-life scenarios. Models that perform well on the test data set tend to be poor in real life where many words have similar pronunciations to keywords, which are confusion words. In this case, to reduce the performance degradation when applying KWS in unmatched scenarios and improve the robustness of KWS, we delivered two systems trained with different augmentation methods. An additional loss is also incorporated to the baseline system in one of the augmentation methods to further improve system's robustness.

\subsection{Masked Audio}
Motivated by approaches in face recognition \cite{wang2020masked}, we applied masking on positive samples and use those samples as part of the negative data in training to improve the robustness of our KWS model. Unlike face recognition, however, the KWS model should yield undetected results when having these masked samples since masked samples are now uncompleted keywords. For each positive sample, we generate 5 corresponding masked samples by replacing 40\%-60\% audio signals with Gaussian white noise. Figure \ref{Truemasked_figure} shows the changes in acoustic features after the replacement.

\subsection{Text-to-speech augmentation}
We obtain synthesized data from a mandarin multi-speaker TTS system \cite{Cai2020}. In this setup, 10k voices from publicly available datasets and internal datasets are collected and used for synthesized. For each speaker, we first extract the speaker embedding with one utterance. Then 2 kinds of synthesized samples are generated by the multi-speaker TTS system conditioned on the speaker embedding: (i) negative samples that do not contains keywords, (ii) adversarial negative samples that have contents close to the keywords. 

\begin{figure*}[ht]
    \begin{center}
        \includegraphics[width=0.9\textwidth]{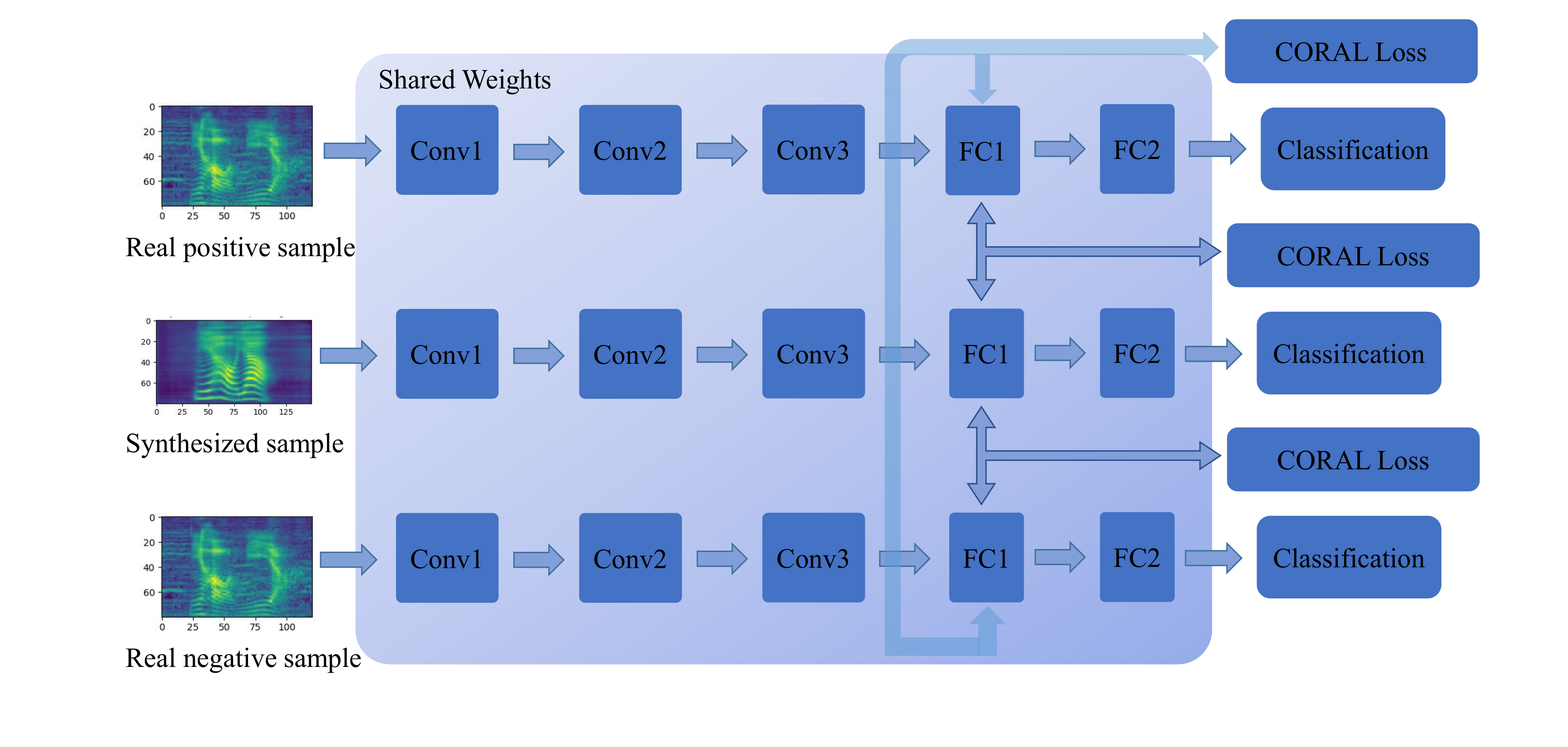}
    \end{center}
    \vspace*{-0.4cm}
    \caption{Framework of the CORAL system.}
    \label{coral_system}
    \vspace*{-0.4cm}
\end{figure*}

\subsection{Correlation Alignment}
We also incorporate the Correlation alignment (CORAL) loss along with TTS augmentation for training to improve the robustness. CORAL is proposed to minimize the difference in second-order statistics between the source and target features \cite{DBLP:journals/corr/SunFS15}. \cite{10.1007/978-3-319-49409-8_35} extended the work to DNN approaches by constructing a differentiable loss function, which can be used to minimize the distance between outputs of the embedding feature layer from different domains. Likewise, \cite{2005.03633} applied the CORAL loss function to incorporate domain knowledge into network training and improved the performance of the keyword classifier on far-field conditions.

Suppose we are given training examples $D_S =$ $\{x_1, x_2,$ $\dots, x_i,$ $\dots, x_{n_S}  \}$ from the source domain and $D_T$ = $\{u_1, u_2,$ $\dots, u_i, \dots, u_{n_T}\}$ from the target domain, where $n_S$ and $n_T$ are the number of samples in source and target domains respectively. Here both $x$ and $u$ are $d$-dimensional vectors. The CORAL loss can then be defined as

\begin{equation}
    L_{CORAL} = \frac{1}{4d^2} ||C_S - C_T ||_F^2
\end{equation}
where $|| . ||_F^2$ denotes the squared matrix Frobenius norm, and $C_S$, $C_T$ denote the covariance matrices of the features in the source domain and the target domain, respectively. The covariance matrices of the source and target data are given by:
\begin{equation}
    C_S = \frac{1}{n_S-1}(D_S^TD_S - \frac{1}{n_S}(\textbf{1}^TD_S)^T(\textbf{1}^TD_S)) 
\end{equation}
\begin{equation}
    C_T = \frac{1}{n_T-1}(D_T^TD_T - \frac{1}{n_T}(\textbf{1}^TD_T)^T(\textbf{1}^TD_T))
\end{equation}
where $\textbf{1}$ is a column vector with all elements equal to 1.

As shown in figure \ref{coral_system}, we calculated the CORAL loss on the hidden feature extracted by CNNs. Data from three individual clusters are paired with each other during training, and the loss is designed as follows:
\begin{scriptsize}
\begin{align}
    &L = L_{ce} + \nonumber \\ &\frac{L_{coral}(C_{real-neg},C_{synt-neg})}{L_{coral}(C_{synt-neg},C_{real-neg}) + L_{coral}(C_{real-pos},C_{real-neg})}  
\end{align}
\end{scriptsize}

where $L_{ce}$ is the cross entropy loss, $C_{real-neg}$, $C_{synt-neg}$ and $C_{real-pos}$ refers to the covariance matrices of embedding features from three clusters (real positive, synthesized negative, real negative). By minimizing the joint loss, the difference of embedding features between synthesized negative samples and real negative samples can be minimized, and the embedding features difference between composite negative samples and real samples can be maximized. 

\section{EXPERIMENT}
\label{sec:pagestyle}
\subsection{Dataset}
Natural speech recorded by native speakers and synthesized speech are both used for training in our experiments.  For natural speech data, the Hi-mia dataset \cite {9054423} is used as the positive training samples. The Hi-mia dataset includes speech data recorded by one close-talking microphone and six 16-channel circular microphone arrays. Each utterance contains content with four Chinese characters ``ni hao, mi ya" (Hello, Mia).  We only use the recordings from the single-channel close-talking microphone. Samples from 300 randomly selected speakers are used as the training set, and samples from 30 speakers are used as the Hi-mia test set. The Aishell-1 \cite {8384449} dataset is used as the negative sample of real speech data. Utterances from 300 speakers are selected for training, and utterances from 30 speakers are used as the Aishell-test set. For synthetic data, we have samples that includes 12 confusion word patterns with 10k different voices, where utterance from 3k speakers are used as the synthetic confusion words (synt-CW) test set. In addition, 188k negative sample audio are synthesized with provided text from  Aishell-2 \cite{1808.10583} (synt-neg). The statistics of the data we used for training and evaluation is shown in table \ref{dataset}, where the term `Real' denotes natural speech, including utterances from Hi-mia dataset (Real Positive) and Aishell-1 (Real Negative).

\begin{table}[h]
    \centering
    \caption{Dataset statistics}        
    \label{dataset}
    \begin{tabular}{cccccc}
        \toprule
        Samples & Train & Test \\
        \midrule
        Real Positive  & 23k    & 2k \\
        Real Negative  & 105k    & 10k \\
        Synthesized Confusion Words      & 90k    & 38k  \\
        Synthetic Negative     & 188k  & - \\
        Masked      &    111k      & - \\
        \bottomrule
    \end{tabular}
    \vspace*{-0.4cm}
\end{table}

\subsection{Experimental setup}
We preprocess the Hi-mia training set by trimming the beginning silence. A speech recognition system trained on the AISHELL-2 dataset is used to perform force alignment on the training set. For each sample, we obtain the start time of pronouncing the word "ni" and use the following 121 frames as the final input, where 121 frames are enough for speaking the keyword ``ni hao, mi ya" (Hello, Mia) according to the alignment information. Our model is trained for 100 epochs with Nesterov momentum Stochastic gradient descent optimizer. The initial learning rate of the optimizer is set to 0.01 and decays when the training loss has not decreased for several rounds. During evaluation, we have a sliding window with a frame length of 121 for each utterance and detect the occurrence of the expected keyword.

Five KWS systems are trained and evaluated regrading different training setups in our experiments:

\begin{enumerate}
\item \textbf{baseline}: use the Real Positive set and the Real Negative set, which are shown in table \ref{dataset}, for training
\item \textbf{real+mask}: use the Real Positive set and the Real Negative set, and the masked samples for training.
\item \textbf{real+synt-CW}: use the Real Positive set and the Real Negative set, and the synt-CW training set for training.
\item \textbf{real+synt-CW+synt-neg}: use the Real Positive set and the Real Negative set, synt-CW training set, and the synt-neg set for training.
\item \textbf{real+synt-CW}: use the Real Positive set and the Real Negative set, and the synt-CW training set for training. CORAL loss is used.
\item \textbf{real+synt-CW+synt-neg+CORAL}:  use the Real Positive set and the Real Negative set, synt-CW training set, and the synt-neg set for training. CORAL loss is incorporated in this setup.
\end{enumerate}
There are two combination sets for evaluation:
\begin{enumerate}
\item \textbf{real}: use the test set from Real Positive set and Real Negative set for evaluation.
\item \textbf{real + synt-CW}: other than the test sets mentioned above, the adversarial synthetic samples are used to evaluation system's robustness on confusion words.
\end{enumerate}

\subsection{Results}

\begin{figure}[h]
    \centering
    \includegraphics[width=0.35\textwidth]{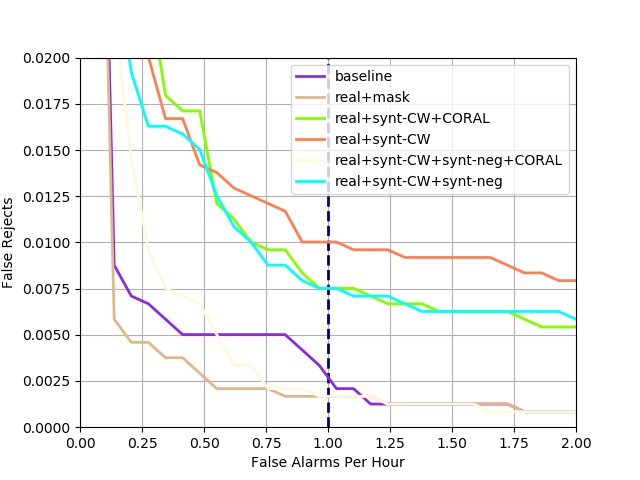}
    \caption{Performances of models on the real test sets}
    \label{fig:result_a}
\end{figure}

\begin{figure}[h]
    \centering
    \includegraphics[width=0.35\textwidth]{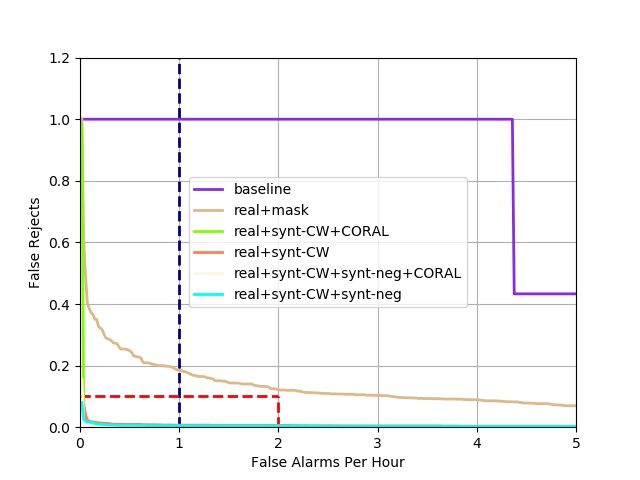}
    \caption{Performances of models on the real+synt-CW test sets}
    \label{fig:result_b}
\end{figure}

\begin{figure}[h]
    \centering
    \includegraphics[width=0.35\textwidth]{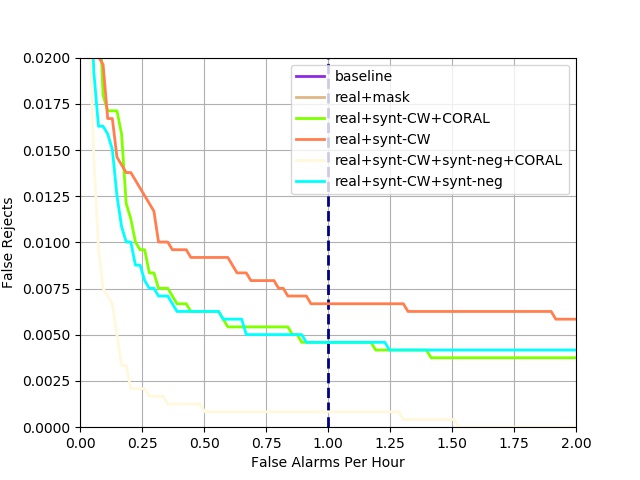}
    \caption{the zoom in view of figure 5}
    \label{fig:result_c}
\end{figure}

\begin{table}[h]
    \footnotesize
    \centering
    \caption{Performances of models trained with different methods on the test sets (the false rejection (FR) rate (\%) under one false alarm (FA) per hour)}        
    \label{result_baseline}
    \begin{tabular}{cccccc}
        \toprule
        Training set & real & real + synt-CW \\
        \midrule
        real(baseline)    & 0.208    & 100.000 \\
        \midrule
        real + masked-pos & \textbf{0.167}    & 18.413 \\
        real + synt-wake     & 1.002    & 0.668  \\
        real + synt-wake + synt-neg    & 0.751  & 0.459 \\
        real + synt-wake + CORAL  & 0.751    & 0.459  \\
        real + synt-wake + synt-neg + CORAL & \textbf{0.167}  & \textbf{0.083} \\
        \bottomrule
    \end{tabular}
    \vspace*{-0.4cm}
\end{table}

Results are shown in figure \ref{fig:result_a}, \ref{fig:result_b} and \ref{fig:result_c}, where figure \ref{fig:result_a} shows the performance of models on real test sets (Hi-mia + Aishell1) without confusing words samples, figure \ref{fig:result_b} shows performance of models on the real + synt-CW test set, and figure \ref{fig:result_c} is the subgraph in the red box from figure \ref{fig:result_b}. We choose the false rejection rate under one false alarm per hour as each model's performance criterion. Table \ref{result_baseline} presents the KWS performance of the sixth models regarding false rejection rate when the false alarm rate per hour is 1.

we can obtain the following observations. First, the baseline system performed well in real test sets without confusing-word samples. However, the baseline system's performance will degrade seriously on confusing-word examples, which is frequently happened in real-life applications. Second, the system trained with masked samples achieves better performance than the baseline system on both the real test set and the confusing-word test set. Third, the confusing-word test set's accuracy has been significantly improved by adding adversarial synthetic training data. However, this method will lead to performance degradation on the real test set (e.g. from 0.208 to 0.751) . In this case, adding synthetic negative samples for training can improve the model's performance in the real test set. Fourth, the training setup incorporated with CORAL loss produces the best results under both the real test set and confusing-word test set. Comparing to the baseline without any augmentation, this augmentation setup achieves better performance on the real testing set, and furthermore, it has shown great robustness on confusion words scenarios as the statistics decrease from 100 to 0.083.

\section{Conclusions}
In this paper, we focus on the task of small-footprint keyword spotting in the confusing-word scenarios and show the effectiveness of incorporating synthesized speech data to train a keyword recognition system. In our daily life, the complex speech environment brings challenges to the recognition system. Confusing words that sound similar to the keywords lead to a degradation in system performance. We adopt two augmentation schemes to enhance the robustness, including masked samples and synthesized samples with the CORAL loss. Experimental results show that our methods can effectively maintain the accuracy on general real test data and at the same time, achieve significant improvement under the condition with confusing word samples.

\vfill\pagebreak
\bibliographystyle{IEEEbib}
\bibliography{strings,refs}

\end{document}